# Combining Ensemble Kalman Filter and Reservoir Computing to predict spatio-temporal chaotic systems from imperfect observations and models

**Futo Tomizawa**[1] **and Yohei Sawada**[1,2,3]

[1]School of Engineering, the University of Tokyo, Tokyo, Japan

[2]Meteorological Research Institute, Japan Meteorological Agency, Tsukuba, Japan

[3]RIKEN Center for Computational Science, Kobe, Japan

**Abstract**

Prediction of spatio-temporal chaotic systems is important in various fields, such as Numerical Weather Prediction (NWP). While data assimilation methods have been applied in NWP, machine learning techniques, such as Reservoir Computing (RC), are recently recognized as promising tools to predict spatio-temporal chaotic systems. However, the sensitivity of the skill of the machine learning based prediction to the imperfectness of observations is unclear. In this study, we evaluate the skill of RC with noisy and sparsely distributed observations. We intensively compare the performances of RC and Local Ensemble Transform Kalman Filter (LETKF) by applying them to the prediction of the Lorenz 96 system. Although RC can successfully predict the Lorenz 96 system if the system is perfectly observed, we find that RC is vulnerable to observation sparsity compared with LETKF. To overcome this limitation of RC, we propose to combine LETKF and RC. In our proposed method, the system is predicted by RC that learned the analysis time series estimated by LETKF. Our proposed method can successfully predict the Lorenz 96 system using noisy and sparsely distributed observations. Most importantly, our method can predict better than LETKF when the process-based model is imperfect.

## 1. Introduction

In Numerical Weather Prediction (NWP), it is required to obtain the optimal estimation of atmospheric state variables by observations and process-based models which are both imperfect. Observations of atmospheric states are sparse and noisy, and numerical models inevitably include biases. In addition, models used in NWP are known to be chaotic, which makes the prediction substantially difficult. To accurately predict the future atmospheric state, it is important to develop methods to predict spatio-tempral chaotic dynamical systems from imperfect observations and models.

Traditionally, data assimilation methods have been widely used in geosciences and NWP systems. Data assimilation is a generic term of approaches to estimate the state from observations and model outputs based on their errors. The state estimated by data assimilation is used as the initial value, and the future state is predicted by models alone. Data assimilation is currently adopted in operational NWP systems. Many data assimilation frameworks have been proposed, e.g. 4D variational methods (4D-VAR; Bannister, 2017), Ensemble Kalman Filter (EnKF; Houtekamer & Zhang, 2016), or their derivatives, and they have been applied to many kinds of weather prediction tasks, such as the prediction of short-term rainfall events (e.g. Sawada et al., 2019; Yokota et al., 2018), and severe storms (e.g. Zhang et al., 2016). Although data assimilation can efficiently estimate the unobservable state variables from noisy observations, the prediction skill is degraded if the model has large biases.

On the other hand, model-free prediction methods based on machine learning have been receiving much attention recently. Many previous studies have successfully applied machine learning to predict chaotic dynamics. Vlachas et al. (2018) successfully applied Long-Short Term Memory (LSTM; Hochreiter & Schmidhuber, 1997) to predict the dynamics of the Lorenz96 model, Kuramoto-Sivashinski Equation, and the barotropic climate model which is a simple atmospheric circulation model. Asanjan et al. (2018) showed that LSTM can accurately predict the future precipitation by learning satellite observation data. Nguyen & Bae (2020) successfully applied LSTM to generate area-averaged precipitation prediction for hydrological forecasting.

In addition to LSTM, Reservoir Computing (RC), which was first introduced by Jaeger & Haas (2004), has been found to be suitable to predict spatio-temporal chaotic systems. Pathak et al. (2017) successfully applied RC to predict the dynamics of Lorenz equation and Kuramoto-Sivashinski Equation. Lu et al. (2017) showed that RC can be used to estimate state variables from sparse observations if the whole system was perfectly observed as training data. Chattopadhyay et al. (2019) revealed that RC can predict the dynamics of the Lorenz 96 model more accurately than LSTM and Artificial Neural Network. In addition to the accuracy, RC also has an advantage in computational costs. RC can learn the dynamics only by training a single matrix just once, while other neural networks have to train numerous parameters and need many iterations (Lu et al., 2017). Thanks to this feature, the computational costs needed to train RC is cheaper than LSTM and Artificial Neural Network.

However, Vlachas et al. (2020) revealed that the prediction accuracy of RC is degraded when all of the state variables cannot be observed. It can be a serious problem since the observation sparsity is often the case in geosciences and the NWP systems. Brajard et al. (2020) pointed out this issue and successfully trained the Convolutional Neural Network with sparse observations, by combining with EnKF. However, their method needs to iterate the data assimilation and training, which is computationally expensive and infeasible toward the real-world problem. Dueben & Bauer (2018) mentioned that the spatio-temporal heterogeneity of observation data made it difficult to train machine learning models, and they suggested to use the model or reanalysis as training data. Weyn et al. (2019) successfully trained machine learning models using the atmospheric reanalysis data.

We aim to propose the novel methodology to predict spatio-temporal chaotic systems from imperfect observations and models. First, we reveal the limitation of the stand-alone use of RC under realistic situations (i.e., imperfect observations and models). Then, we propose a new method to maximize the potential of RC to predict chaotic systems from imperfect models and observations, which is even computationally feasible. As Dueben & Bauer (2018) proposed, we make RC learn the analysis data series generated by a data assimilation method. Our new method can accurately predict from imperfect observations. Most importantly, we found that our proposed method is more robust to model biases than the stand-alone use of data assimilation methods.

## 2. Methods

### 2.1 Lorenz 96 model and OSSE

We used a low dimensional spatio-temporal chaotic model, the Lorenz 96 model (L96), to perform experiments with various parameter settings. L96 is a model introduced by Lorenz & Emanuel (1998) and has been commonly used in data assimilation studies (e.g. Kotsuki et al., 2017; Miyoshi, 2005; Penny, 2014; Raboudi et al., 2018). L96 is recognized as a good testbed for the operational NWP problems (Penny, 2014).

In this model, we consider a ring structured and $m$ dimensional discrete state space $x_1, x_2, \ldots, x_m$ (that is, $x_m$ is adjacent to $x_1$), and define the system dynamics as follows:

$$\frac{dx_i}{dt} = (x_{i+1} - x_{i-2}) x_{i-1} - x_i + F \quad (1)$$

where $F$ stands for the force parameter. Each term of this equation corresponds to advection, damping and forcing respectively. It is known that various settings of state dimension $m$, forcing term $F$ and initial values result in chaotic solutions. The time width $\Delta t = 0.2$ corresponds to one day in real atmospheric motion from the view of Lyapunov time (Miyoshi, 2005).

As we use this conceptual model, we cannot obtain any observational data or "true" phenomena that correspond to the model. Instead, we adopted Observing System Simulation Experiment (OSSE). We first prepared a time series by integrating equation (1) and regarded it as the "true" dynamics (called Nature Run). Observation data can be calculated from this time series adding some perturbation:

$$\boldsymbol{y^o} = \boldsymbol{Hx} + \boldsymbol{\epsilon} \quad (2)$$

where $\boldsymbol{y^o} \in \mathbb{R}^h$ is the observation value, $\boldsymbol{H}$ is the $m \times h$ observation matrix, $\boldsymbol{\epsilon} \in \mathbb{R}^h$ is the observational error whose each element is independent and identically distributed on Gaussian distribution $N(0, e)$ for observation error $e$.

In each experiment, the form of L96 used to generate Nature Run is unknown, and the model used to make prediction can be different from that for Nature Run. The difference between the model used for Nature Run and that used for prediction corresponds to the model's bias in the context of NWP.

### 2.2 Local Ensemble Transform Kalman Filter

We used the Local Ensemble Transform Kalman Filter (LETKF, Hunt et al., 2007) as the data assimilation method in this study. LETKF is one of the ensemble-based data assimilation methods, which is considered to be applicable to the NWP problems in many previous studies (Sawada et al., 2019; Yokota et al., 2018). LETKF is also used for the operational NWP in some countries (e.g. Schraff et al., 2016).

LETKF and the family of ensemble Kalman filters have two steps; forecast and analysis. The forecast step makes the prediction from the analysis variables of current time to the time when the next observation is obtained (this time width is called "assimilation window"). Considering the stochastic error in the model, system dynamics can be represented as follows (hereafter the subscript $k$ stands for the variable at time $k$, and the time width of $k$ corresponds to the assimilation window):

$$x_k^f = \mathcal{M}(x_{k-1}^a) + \eta_k, \qquad \eta_k \sim N(0, Q) \tag{3}$$

where $x_k^f \in \mathbb{R}^m$ is the forecast variables, $x_{k-1}^a \in \mathbb{R}^m$ is the analysis variables, $\mathcal{M}: \mathbb{R}^m \to \mathbb{R}^m$ is the model dynamics operator, $\eta \in \mathbb{R}^m$ is the stochastic error and $N(0, Q)$ means the Gaussian distribution with mean $0$ and $n \times n$ covariance matrix $Q$. Using the computed state vector $x_k^f$, observation variables can be estimated as follows:

$$y_k^f = \mathcal{H}(x_k^f) + \epsilon_k, \qquad \epsilon_k \sim N(0, R) \tag{4}$$

where $y^f \in \mathbb{R}^h$ is the estimated observation value, $\mathcal{H}: \mathbb{R}^m \to \mathbb{R}^h$ is the observation operator and $\epsilon \in \mathbb{R}^h$ is the observation error extracted from $N(0, R)$. Since the error in the model is assumed to follow the Gaussian distribution, forecasted state $x^f$ can also be considered as a random variable from the Gaussian distribution if $\mathcal{M}$ is linear. In this situation, the probability distribution of $x^f$ (and also $x^a$) can be parametrized by mean $\overline{x^f}$ ($\overline{x_k^a}$,) and covariance matrix $P^f$ ($P_k^a$). Their temporal evolution can be calculated based on equation (3) as follows:

$$\overline{x_{k+1}^f} = M\overline{x_k^a}, \qquad P_k^f = MP_k^a M^T + Q \tag{5}$$

where $M$ is the $m \times m$ matrix representation of $\mathcal{M}$. Hereafter the means of $x^f$ and $x^a$ are expressed without overlines for convenience.

Next, in the analysis step, this forecast state is updated using actual observation $y_k^o$. $x_k^a$ and $P_k^a$ are generated as follows:

$$x_k^a = x_k^f + K_k(y^o - Hx_k^f), \qquad P_k^a = (I - K_k H)P_k^f$$

$$K_k = P_k^f H^T (HP_k^f H^T + R)^{-1} \tag{6}$$

where $H$ is the linear observation operator of equation (4). This method is called Kalman Filter. Kalman Filter is a good approximation when the dynamics is linear. However, it is difficult to apply it to nonlinear and large problems. If either the model operator $\mathcal{M}$ or observation operator $\mathcal{H}$ is nonlinear, we cannot directly use this method. If the state space dimension $n$ is high, it is difficult to keep $n \times n$ covariance matrix $P$ on the memory.

One of the methods that solve these problems is EnKF. EnKF uses an ensemble of state variables to represent the probability distribution. The forecast step of equation (5) then becomes as follows:

$$x_k^{f,(i)} = \mathcal{M}(x_{k-1}^{a,(i)}), \qquad P_k^f = \frac{1}{N_e - 1} X_k^f (X_k^f)^T \tag{7}$$

where $x_k^{f,(i)}$ is the $i$th ensemble member of forecast value at time $k$, $N_e$ is the number of ensemble members and $X_k^f$ is the matrix whose $i$th column is the deviation of the $i$th ensemble member from the ensemble mean.

The analysis step of EnKF has some variants including LETKF. LETKF first determines the mean and covariance of the analysis ensemble, $\overline{x_k^a}$ and $P_k^a$, and then computes the analysis ensemble. As the derivation of equation (6), we get $\overline{x_k^a}$ and $P_k^a$ from forecast ensemble as follows:

$$\overline{w_k^a} = \widetilde{P}_k^a \left(HX_k^f\right)^T R^{-1} \left(y^o - H\overline{x_k^f}\right)$$
$$\widetilde{P}_f^a = \left[(k-1)I + \left(HX_k^f\right)^T R^{-1} HX_k^f\right]^{-1} \quad (8)$$
$$\overline{x_k^a} = \overline{x_k^f} + X_k^f \overline{w_k^a}$$
$$P_k^a = X_k^f \widetilde{P}_k^a \left(X_k^f\right)^T$$

where $w_k^a, \widetilde{P}_f^a$ stands for the mean and covariance of the analysis ensemble calculated in the ensemble subspace. As equation (7), we can consider the analysis covariance as the product of the analysis ensemble matrix:

$$P_k^a = \frac{1}{N_e - 1} X_k^a (X_k^a)^T \quad (9)$$

where $X_k^a$ is the matrix whose $i$th column is the variation of the $i$th ensemble member from the mean for the analysis ensemble. Therefore, decomposing $\widetilde{P}_k^a$ of equation (8) into square root, we can get each analysis ensemble member at time $k$ as follows:

$$W_k^a (W_k^a)^T = \widetilde{P}_k^a, \quad x_k^a = \sqrt{N_e - 1} \, X_k^f w_k^a \quad (10)$$

where $w_k^a$ is the $i$th column of $W_k^a$ in the first equation. A covariance inflation parameter is multiplied to take measures for the tendency of data assimilation to underestimate the uncertainty of state estimate. See Hunt et al. (2007) for more detailed derivation. Now, we can return to the equation (7) and iterate forecast and analysis step.

As in the real application, we consider the situation that the observations are not available in the prediction period. Predictions are made by the model alone, using the latest analysis state variables as the initial condition. This way of making prediction is called "Extended Forecast", and we call this prediction "LETKF-Ext" in this study.

## 2.3 Reservoir Computing

We use Reservoir Computing (RC) as the machine learning framework. RC is a type of Recurrent Neural Network, which has a single hidden layer called reservoir. Figure 1 shows the architecture. As mentioned in the Section 1, the previous works have shown that RC can predict the dynamics of spatio-temporal chaotic systems.

The state of the reservoir layer at timestep $k$ is represented as a vector $r_k \in \mathbb{R}^{D_r}$, which evolves given the input vector $u_k \in \mathbb{R}^m$ as follows:

$$r_{k+1} = \tanh[Ar_k + W_{in} u_k] \quad (11)$$

where $W_{in}$ is the $D_r \times M$ input matrix which maps the input vector to the reservoir space, and $A$ is the $D_r \times D_r$ adjacency matrix of the reservoir which determines the reservoir dynamics. $W_{in}$ should be determined to have only one nonzero component in each row, and each nonzero component is extracted from uniform distribution of $[-a, a]$ for some parameter $a$. $A$ has a proportion of $d$ nonzero components with random values from uniform distribution, and it is normalized to have the maximum eigenvalue $\rho$. The reservoir size $D_r$ should be determined based on the size of the state space. From the reservoir state, we can compute the output vector $v$ as follows:

$$v_k = W_{out} f(r_k) \quad (12)$$

where $W_{out}$ is the $M \times D_r$ output matrix which maps the reservoir state to the state space, and $f: \mathbb{R}^{D_r} \to \mathbb{R}^{D_r}$ is the operator for nonlinear transformation. The nonlinear transformation is essential for the accurate prediction (Chattopadhyay et al., 2019). It is important that $A$ and $W_{in}$ are fixed and only $W_{out}$ will be trained. Therefore, the computational cost required to train RC is small and it is an outstanding advantage of RC compared to the other neural network frameworks.

In the training phase, we set the switch in the Figure 1 to the training configuration. Given a training data series $\{u_0, u_1, \ldots, u_n\}$, we can generate the reservoir state series $\{r_1, r_2, \ldots, r_{n+1}\}$ by equation (11) . By using the training data and reservoir state series, we can determine the $W_{out}$ matrix by ridge regression. We minimize the following square error function with respect to $W_{out}$:

$$\sum_{i=1}^{n} \|u_k - W_{out} f(r_k)\|^2 + \beta \cdot trace(W_{out} W_{out}^T) \qquad (13)$$

where $\|x\| = x^T x$ and $\beta$ is the ridge regression parameter (normally a small positive number). The optimal value can be determined analytically as follows:

$$W_{out} = UR^T (RR^T + \beta I)^{-1} \qquad (14)$$

where $I$ is the $D_r \times D_r$ identity matrix and $R, U$ are the matrix whose $kth$ column is the vector $f(r_k), u_k$, respectively.

Then, we can shift to the predicting phase. Before we predict with the network, we first need to "spin up" the reservoir state. The spin up process was done by giving the time series before the initial value $\{u_{-k}, u_{-k+1}, \ldots, u_{-1}\}$ to the network and calculate the reservoir state right before the beginning of the prediction via equation (11). After that, the output layer is connected to the input layer, and the network becomes recursive. In this configuration, the output value $v_k$ of equation (12) is used as the next input value $u_k$ of equation (11). Once we give the initial value $u_0$, the network will iterate equation (11) and (12) spontaneously, and the prediction will be yielded.

Considering the real application, it is natural to assume that the observation data can only be used as the training data and the initial value for the RC prediction. In this paper we call this type of prediction "RC-Obs".

**2.4 Combination of RC and LETKF**

As discussed so far and we will quantitatively discuss in the section 4, LETKF-Ext and RC-Obs have contrasting advantages and disadvantages. LETKF-Ext can accurately predict even if the observation is noisy and/or sparsely distributed, while RC-Obs is vulnerable to the imperfectness in observation. On the other hand, LETKF-Ext can be strongly affected by the model biases since the prediction of LETKF-Ext depends only on the model after obtaining the initial condition, while RC-Obs has no dependence to the accuracy of the model as it only uses the observation data for training and prediction.

Therefore, the combination of LETKF and RC has a potential to push the limit of these two individual prediction methods and realize accurate and robust prediction. The weakness of RC-Obs is that we can only use the observational data directly, which is inevitably sparse in the real application, although RC is vulnerable to this imperfectness. In our proposed method, we make RC learn the analysis time series generated by LETKF instead of directly learning observation data. Since LETKF's analysis variables are of full grid, it is expected that we can efficiently train RC in our proposed method. We call the prediction by this method "RC-Anl".

Our proposed combination method is expected to predict more accurately than RC-Obs since the training data always exist in all the grid points, even if the observation is sparse. Also, especially if the model is substantially biased, the analysis time series generated by LETKF is more accurate than the model output itself. It means that RC-Anl is expected to be able to predict more accurately than LETKF-Ext.

## 3. Experiment Design

To generate the Nature Run, L96 with $m = 8$, $F = 8$ was used, and it was numerically integrated by 4th order Runge-Kutta method by time width $\Delta t = 0.005$. Before calculating the Nature Run, the L96 equation was integrated for 1440000 timesteps for spin up. In the following experiment, $F$ term in the model was changed to represent the model bias.

The setting for LETKF was based on Miyoshi & Yamane (2007). In equation (8), each row of observation covariance $\boldsymbol{R}$ were divided by the value $w$ calculated as follows:

$$w(r) = \exp\left(\frac{r^2}{18}\right) \tag{15}$$

where $r$ is the distance between each observation point and each analyzed point. The shape of equation (8) differs by the analyzed grid points, so each row of $w_k^a$ and $\tilde{P}_k^a$ should be calculated separately. In equation (10), a "covariance inflation factor", which was set to 1.05 in our study, was multiplied to $\tilde{P}_k^a$ in each iteration. Ensemble size $N_e$ was set to 20.

The configuration of RC used in this study was similar to Chattopadhyay et al. (2019), but was slightly modified. Parameter settings used in the RC experiments are shown in Table 1. The nonlinear transformation function for the output layer in equation (12) is as follows:

$$f(r_i) = \begin{cases} r_i & (i \text{ is } odd) \\ r_{i-1} \times r_{i-2} & (i \text{ is } even) \end{cases} \tag{16}$$

where $r_i$ is the $i$th element of $\boldsymbol{r}$. In the prediction phase, we used the data for 100 timesteps before the prediction initial time for the reservoir spin up.

We implemented numerical experiments to investigate the performance of RC-Obs, LETKF-Ext and RC-Anl to predict L96 dynamics. First, we evaluated the performance of RC-Obs by comparing with LETKF-Ext under perfect observations (all the grid points are observed with no error) and quantified the effect of the observation imperfectness (i.e. observation error and spatio-temporal sparsity), to investigate the prediction skill of the stand-alone use of RC and LETKF. Second, we evaluated the performance of RC-Anl. We investigated the performance of RC-Anl and

LETKF-Ext as the functions of the observation density and model biases. Three prediction frameworks are summarized in Table 2.

In each experiment, we prepared 200000 timesteps of Nature Run. The first 100000 timesteps were used for the training of RC or for the spinning up of LETKF, and the rest of them were used for the evaluation of each method. Every prediction was repeated 100 times to avoid the effect of the heterogeneity of data. For the LETKF-Ext prediction, the analysis time series of all the evaluation data was firstly generated. Then, the analysis variables for one every 1000 timestep was taken as the initial conditions and total 100 prediction runs were performed. For the RC-Obs prediction, evaluation data were equally divided into 100 sets and the prediction was identically done for each set. For the RC-Anl prediction, the analysis time series of training data were used for training, and the prediction was performed using the same initial condition as LETKF-Ext. Each prediction set of LETKF-Ext, RC-Obs, and RC-Anl corresponds to the same time range.

The prediction accuracy of each method was evaluated by taking the average of RMSE of 100 sets for each timestep. We call this metric mean RMSE ($mRMSE$), and can be represented as follows:

$$mRMSE(t) = \frac{1}{100}\sum_{i=1}^{100}\sqrt{\frac{1}{m}\sum_{j=1}^{m}\left(u_j^{(i)}(t) - x_j^{(i)}(t)\right)^2} \tag{17}$$

where $t$ is the number of the steps elapsed from the prediction initial time, $x_j^{(i)}(t)$ is the $j$th nodal value of the $i$th prediction set at time $t$ and $u_j^{(i)}(t)$ is the corresponding value of Nature Run. Using this metric, we can see how the prediction accuracy is degraded as time elapses from initial time.

## 4. Results

Figure 2 shows the Hovmöller diagram of Nature Run, LETKF-Ext, and RC-Obs. Figure 2 also shows the difference between prediction and Nature Run, as well as the actual prediction result so that we can see how long we can keep the prediction accurate. The model and observation used for each prediction was perfect, that is, the model was the same as the one for Nature Run, and the observation was available for all the grid point and every timestep, with observation error $e = 0.01$ (if it is set to 0, LETKF does not work). Although both predictions are accurate in the short lead time, LETKF-Ext can accurately predict the state variables for the longer lead time than RC-Obs. If we have perfect model and observations, the prediction skill of LETKF-Ext is better than RC-obs.

Figure 3 shows the time variation of the $mRMSE$ (see equation (17)) of LETKF-Ext and RC-Obs. This figure clarifies the superiority of LETKF-Ext. The $mRMSE$ of LETKF-Ext was less than that of RC-Obs at all timesteps.

Next, we evaluated the sensitivity of the prediction skill of both LETKF-Ext and RC-Obs to the imperfectness of the observations. Figure S1 and Figure S2 show the effect of the observation error and frequency on the prediction skill, respectively. Both methods showed a similar level of robustness for the change of the observation frequency and the observation error.

However, if we reduce the number of the observed grid points, the prediction accuracy of RC-Obs becomes catastrophically worse while LETKF-Ext is robust to the reduced number of the observed grid points. Figures 4a and 4b show the sensitivity of the prediction accuracy of LETKF-Ext and RC-Obs, respectively, to the number of observed grid points. Even though we can observe a small part of the system, the accuracy of LETKF-Ext changed only slightly. On the other hand, the accuracy of RC-Obs gets substantially worse when we remove a single observed grid point. As assumed in the section 2.4, we verified that RC-Obs is significantly sensitive to the observation sparsity.

We tested the prediction skill of our newly proposed method, RC-Anl, under imperfect models and sparse observations. Here, we used the observation error $e = 1.0$. Figure 5 shows the change of the $mRMSE$ time series of RC-Anl with the different number of observed grid points. It indicates that the vulnerability of the prediction accuracy to the change of the number of observed grid points, which is found in RC-Obs, no longer exists in RC-Anl. Although the prediction accuracy is lower than LETKF-Ext (Figure 4a), our new method indicates a robustness to the observation sparsity and overcomes the limitation of the stand-alone RC.

Moreover, when the model used in LETKF is biased, RC-Anl outperforms LETKF-Ext. Figure 6 shows the change of the $mRMSE$ time series when changing the model biases. The number of the observed points was set to 4. The $F$ term in equation (1) was changed from the true value 8 (the $F$ value of the model for Nature Run) as the model bias, and the accuracy of LETKF-Ext and RC-Anl is plotted. The accuracy of LETKF-Ext was slightly better than that of RC-Anl when the model was not biased ($F = 8$; green line). However, when the bias is large (e.g. $F = 10$; yellow line), RC-Anl showed the better prediction accuracy.

We confirmed this result by comparing the $mRMSE$ value of RC-Anl and LETKF-Ext at the specific forecast lead-time. Figure 7 shows the value of $mRMSE(80)$ (see equation (17)) as the function of the value of the $F$ term. Both two lines that shows the skill of RC-Anl (blue) and LETKF-Ext (red) are convex downward and have a minimum at $F = 8$, meaning that the accuracy of both prediction methods are the best when the model is not biased. In addition, as long as $F$ value is in the interval $[7.5, 8.5]$, LETKF-Ext has the better accuracy than RC-Anl. However, if the model bias become larger than that, RC-Anl becomes more accurate than LETKF-Ext. As the bias increases, the difference between the $mRMSE(80)$ of two methods becomes larger, and the superiority of RC-Anl becomes more obvious. We found that RC-Anl can predict more accurately than LETKF-Ext when the model is biased.

We also checked the robustness for the training data size. Figure S3 shows the change of the accuracy of RC-Anl by changing the size of training data from 100000 to 1000 timesteps. We confirmed that the prediction accuracy did not change until the size was reduced to 25000 timesteps. Although we have used a large size of training data (100000 timesteps; 68 model years) so far, the results are robust to the reduction of the size of the training data.

## 5. Discussion

By comparing the prediction skill of RC-Obs and LETKF-Ext, we confirmed that RC-Obs can predict with accuracy comparable to LETKF-Ext, if we have perfect observations. This result is consistent with Chattopadhyay et al. (2019), Pathak et al. (2017) or P. R. Vlachas et al. (2020), and we can expect that RC has a potential to predict various kinds of spatio-temporal chaotic systems.

However, Vlachas et al. (2020) revealed that the prediction accuracy of RC is substantially degraded when the observed grid points are reduced, compared to other machine learning techniques such as LSTM. Our result is consistent with their study, and we found that the prediction accuracy of RC-Obs was significantly degraded by just removing one observation grid point. In contrast, Chattopadhyay et al. (2019) showed that RC can predict the multi-scale chaotic system correctly even though only the largest scale dynamics is observed. Comparing these results, we can suggest that the states in the scale of dominant dynamics should be observed almost perfectly to accurately predict the future state by RC.

Therefore, when we use RC to predict spatio-temporal chaotic systems with sparse observation data, we need to interpolate them to generate the appropriate training data. However, the interpolated data inevitably includes errors even if the observation data itself has no error, so it should be verified that RC can predict accurately by training data with some errors. Previous works such as Chattopadhyay et al., 2019, Pathak et al., 2017, or P. R. Vlachas et al., 2020 have not considered the impact of error in the training data. We found that the prediction accuracy of RC degrades as the error in training data grows, but the degradation rate is not so large (if all the training data of all the grid points are obtained). We can expect from this result that RC trained with the interpolated observation data can predict accurately to some extent, but the interpolated data should be as accurate as possible.

In this study, LETKF was used to prepare the training data for RC, since LETKF can interpolate the observations and reduce their error at the same time. We showed that our proposed approach correctly works. Brajard et al. (2020) also made Convolutional Neural Network (CNN) learn the dynamics from sparse observation data and successfully predict the dynamics of the L96 model. However, as mentioned in the introduction section, Brajard et al. (2020) needed to iterate the learning and data assimilation until they converge, because it replaced the model used in data assimilation with CNN. Although their model-free method has an advantage that it was not affected by the process-based model's reproducibility of the phenomena, it is computationally expensive and probably infeasible in many real-world problems. Contrary, we need to train RC just one time, because we use the process-based model (i.e. data assimilation method) to prepare the training data. We overcome the problem of computational feasibility. Note also that the computational cost to train RC is much cheaper than the other neural networks.

The good performance of our proposed method supports the suggestion of Dueben & Bauer (2018), in which machine learning should be applied to the analysis data generated by data assimilation methods as the first step of the application of machine learning to weather prediction. As Weyn et al. (2019) did, we successfully trained the machine learning model with the analysis data.

Most importantly, we also found that the prediction by RC-Anl is more robust to the model biases than the extended forecast by LETKF (i.e. LETKF-Ext). This result suggests that our method can be beneficial in various real problems, as the model in real applications inevitably contains some biases. Pathak, Wikner, et al. (2018) developed the hybrid prediction system of RC and a biased model. Although Pathak, Wikner et al. (2018) successfully predicted the spatio-temporal chaotic systems using the biased models, they needed perfect observations to train their RC. The advantage of our proposed method is that we allow both models and observation networks to be imperfect.

Our study was implemented with the 8-dimensional L96 system, and it is unclear whether our proposed method is applicable to other spatio-temporal chaotic systems with larger state spaces, including the real NWP models. However, in previous works, RC has been successfully applied to many other large chaotic systems. Especially, Pathak, Hunt, et al. (2018) indicated that RC can be applied to predict the dynamics of substantially high dimensional Kuramoto-Sivashinski equation using the "reservoir parallelization". They divided the state space to some local groups and used different reservoirs for each local group. As we did not change the RC architecture itself, our method also has a potential to predict other high dimensional spatio-temporal chaotic systems by adopting this parallelization strategy.

In NWP problems, it is often the case that homogenous observation data of high resolution are not available over a wide range of time and space, which can be an obstacle to applying machine learning to NWP tasks (Dueben & Bauer, 2018). We revealed that RC is robust for the temporal sparsity of observations, and RC can be trained with relatively small training data sets. These results imply that our proposed method can be applicable to various realistic problems.

## 6. Conclusion

The prediction skills of the extended forecast with LETKF (LETKF-Ext), RC that learned the observation data (RC-Obs), and RC that learned the LETKF analysis data (RC-Anl) were evaluated under imperfect models and observations, using the Lorenz 96 model. We found that the prediction by RC-Obs is substantially vulnerable to the sparsity of the observation network. Our proposed method, RC-Anl, can overcome this vulnerability. In addition, RC-Anl could predict more accurately than LETKF-Ext when the process-based model is biased. Our new method is robust to the imperfectness of both models and observations so that it is feasible to apply it to the real NWP problem. Further studies on more complicated models or operational atmospheric models are expected.

**Code Availability**

The source code for RC and Lorenz96 model is available at: https://doi.org/10.5281/zenodo.3907291, and for LETKF at: https://github.com/takemasa-miyoshi/letkf

**Acknowledgement**

This work was supported by the Japan Society for the Promotion of Science KAKENHI grant JP17K18352 and JP18H03800, the JAXA grant ER2GWF102, and the JST AIP Grant Number JPMJCR19U2.

# References

A. Asanjan, A., Yang, T., Hsu, K., Sorooshian, S., Lin, J. and Peng, Q.: Short-Term Precipitation Forecast Based on the PERSIANN System and LSTM Recurrent Neural Networks, J. Geophys. Res. Atmos., 123(22), 12,543-12,563, doi:10.1029/2018JD028375, 2018.

Bannister, R. N.: A review of operational methods of variational and ensemble-variational data assimilation, Q. J. R. Meteorol. Soc., 143(703), 607–633, doi:10.1002/qj.2982, 2017.

Brajard, J., Carrassi, A., Bocquet, M. and Bertino, L.: Combining data assimilation and machine learning to emulate a dynamical model from sparse and noisy observations : a case study with the Lorenz 96 model, arXiv, doi:arXiv:2001.01520v1, 2020a.

Brajard, J., Carrassi, A., Bocquet, M. and Bertino, L.: Combining data assimilation and machine learning to emulate a dynamical model from sparse and noisy observations: a case study with the Lorenz 96 model, J. Comput. Sci., 1–18, doi:10.1016/j.jocs.2020.101171, 2020b.

Chattopadhyay, A., Hassanzadeh, P. and Subramanian, D.: Data-driven prediction of a multi-scale Lorenz 96 chaotic system using deep learning methods: Reservoir computing, ANN, and RNN-LSTM. [online] Available from: https://doi.org/10.31223/osf.io/fbxns, 2019.

Dueben, P. D. and Bauer, P.: Challenges and design choices for global weather and climate models based on machine learning, Geosci. Model Dev., 11(10), 3999–4009, doi:10.5194/gmd-11-3999-2018, 2018.

Hochreiter, S. and Schmidhuber, J.: Long Short-Term Memory, Neural Comput., 9(8), 1735–1780, doi:10.1162/neco.1997.9.8.1735, 1997.

Houtekamer, P. L. and Zhang, F.: Review of the ensemble Kalman filter for atmospheric data assimilation, Mon. Weather Rev., 144(12), 4489–4532, doi:10.1175/MWR-D-15-0440.1, 2016.

Hunt, B. R., Kostelich, E. J. and Szunyogh, I.: Efficient data assimilation for spatiotemporal chaos: A local ensemble transform Kalman filter, Phys. D Nonlinear Phenom., 230(1–2), 112–126, doi:10.1016/j.physd.2006.11.008, 2007.

Jaeger, H. and Haas, H.: Harnessing Nonlinearity: Predicting Chaotic Systems and Saving Energy in Wireless Communication, Science (80-. )., 304(5667), 78–80, 2004.

Kotsuki, S., Greybush, S. J. and Miyoshi, T.: Can we optimize the assimilation order in the serial ensemble Kalman filter? A study with the Lorenz-96 model, Mon. Weather Rev., 145(12), 4977–4995, doi:10.1175/MWR-D-17-0094.1, 2017.

Lorenz, E. N. and Emanuel, K. A.: Optimal sites for supplementary weather observations: Simulation with a small model, J. Atmos. Sci., 55(3), 399–414, doi:10.1175/1520-0469(1998)055<0399:OSFSWO>2.0.CO;2, 1998.

Lu, Z., Pathak, J., Hunt, B., Girvan, M., Brockett, R. and Ott, E.: Reservoir observers: Model-free inference of unmeasured variables in chaotic systems, Chaos, 27, 041102, doi:10.1063/1.4979665, 2017.

Miyoshi, T.: ENSEMBLE KALMAN FILTER EXPERIMENTS WITH A PRIMITIVE-EQUATION GLOBAL MODEL, Ph.D. Diss. Univ. Maryland, Coll. Park, (2002), 197, 2005.

Miyoshi, T. and Yamane, S.: Local ensemble transform Kalman filtering with an AGCM at a T159/L48 resolution, Mon. Weather Rev., 135(11), 3841–3861, doi:10.1175/2007MWR1873.1, 2007.

Nguyen, D. H. and Bae, D. H.: Correcting mean areal precipitation forecasts to improve urban flooding predictions by using long short-term memory network, J. Hydrol., 584(February), 124710, doi:10.1016/j.jhydrol.2020.124710, 2020.

Pathak, J., Lu, Z., Hunt, B. R., Girvan, M. and Ott, E.: Using machine learning to replicate chaotic attractors and calculate Lyapunov exponents from data, Chaos, 27, 121102, doi:10.1063/1.5010300, 2017.

Pathak, J., Wikner, A., Fussell, R., Chandra, S., Hunt, B. R., Girvan, M. and Ott, E.: Hybrid forecasting of chaotic processes: Using machine learning in conjunction with a knowledge-based model, Chaos, 28(4), doi:10.1063/1.5028373, 2018a.

Pathak, J., Hunt, B., Girvan, M., Lu, Z. and Ott, E.: Model-Free Prediction of Large Spatiotemporally Chaotic Systems from Data: A Reservoir Computing Approach, Phys. Rev. Lett., 120, 024102, doi:10.1103/PhysRevLett.120.024102, 2018b.

Penny, S. G.: The hybrid local ensemble transform Kalman filter, Mon. Weather Rev., 142(6), 2139–2149, doi:10.1175/MWR-D-13-00131.1, 2014.

Raboudi, N. F., Ait-El-Fquih, B. and Hoteit, I.: Ensemble Kalman filtering with one-step-ahead smoothing, Mon. Weather Rev., 146(2), 561–581, doi:10.1175/MWR-D-17-0175.1, 2018.

Sawada, Y., Okamoto, K., Kunii, M. and Miyoshi, T.: Assimilating Every-10-minute Himawari-8 Infrared Radiances to Improve Convective Predictability, J. Geophys. Res. Atmos., 124(5), 2546–2561, doi:10.1029/2018JD029643, 2019.

Schraff, C., Reich, H., Rhodin, A., Schomburg, A., Stephan, K., Periáñez, A. and Potthast, R.: Kilometre-scale ensemble data assimilation for the COSMO model (KENDA), Q. J. R. Meteorol. Soc., 142(696), 1453–1472, doi:10.1002/qj.2748, 2016.

Vlachas, P. R., Byeon, W., Wan, Z. Y., Sapsis, T. P. and Koumoutsakos, P.: Data-driven forecasting of high-dimensional chaotic systems with long short-Term memory networks, Proc. R. Soc. A Math. Phys. Eng. Sci., 474(2213), doi:10.1098/rspa.2017.0844, 2018.

Vlachas, P. R., Pathak, J., Hunt, B. R., Sapsis, T. P., Girvan, M., Ott, E. and Koumoutsakos, P.: Backpropagation algorithms and Reservoir Computing in Recurrent Neural Networks for the forecasting of complex spatiotemporal dynamics, Neural Networks, 126, 191–217, doi:10.1016/j.neunet.2020.02.016, 2020.

Weyn, J. A., Durran, D. R. and Caruana, R.: Can Machines Learn to Predict Weather? Using Deep Learning to Predict Gridded 500-hPa Geopotential Height From Historical Weather Data, J. Adv. Model. Earth Syst., 11(8), 2680–2693, doi:10.1029/2019MS001705, 2019.

Yokota, S., Seko, H., Kunii, M., Yamauchi, H. and Sato, E.: Improving Short-Term Rainfall Forecasts by Assimilating Weather Radar Reflectivity Using Additive Ensemble Perturbations, J. Geophys. Res. Atmos., 123(17), 9047–9062, doi:10.1029/2018JD028723, 2018.

Zhang, F., Minamide, M. and Clothiaux, E. E.: Potential impacts of assimilating all-sky infrared satellite radiances from GOES-R on convection-permitting analysis and prediction of tropical cyclones, Geophys. Res. Lett., 43(6), 2954–2963, doi:10.1002/2016GL068468, 2016.

Table 1. Parameter values of RC used in each experiment

| Parameter | Description | Value |
|---|---|---|
| $D_r$ | reservoir size | 5000 |
| $a$ | Input matrix scale | 0.5 |
| $d$ | adjacency matrix density | 0.0006 |
| $\rho$ | adjacency matrix spectral radius | 0.1 |
| $\beta$ | ridge regression parameter | 0.0001 |

Table 2. Summary of three prediction frameworks

| Name | Initial Value | Model for prediction |
|---|---|---|
| LETKF-Ext | LETKF analysis | the model used in LETKF |
| RC-Obs | observation | RC trained with observation |
| RC-Anl | LETKF analysis | RC trained with LETKF analysis |

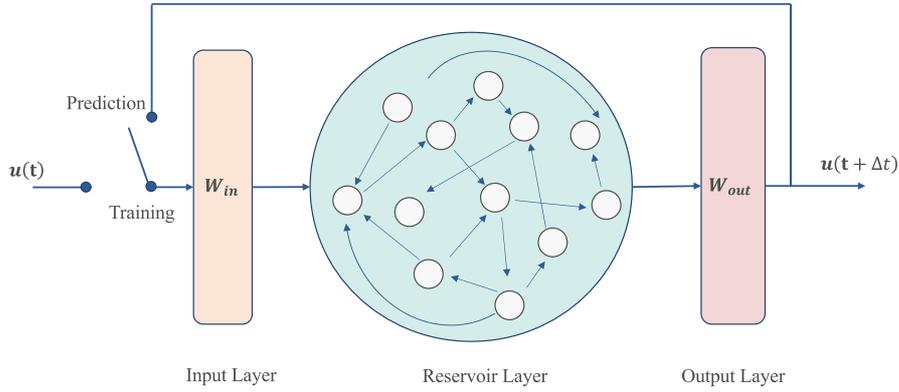

**Figure 1**. The conceptual diagram of reservoir computing architecture. The network consists of an input layer, a hidden layer called reservoir, and an output layer.

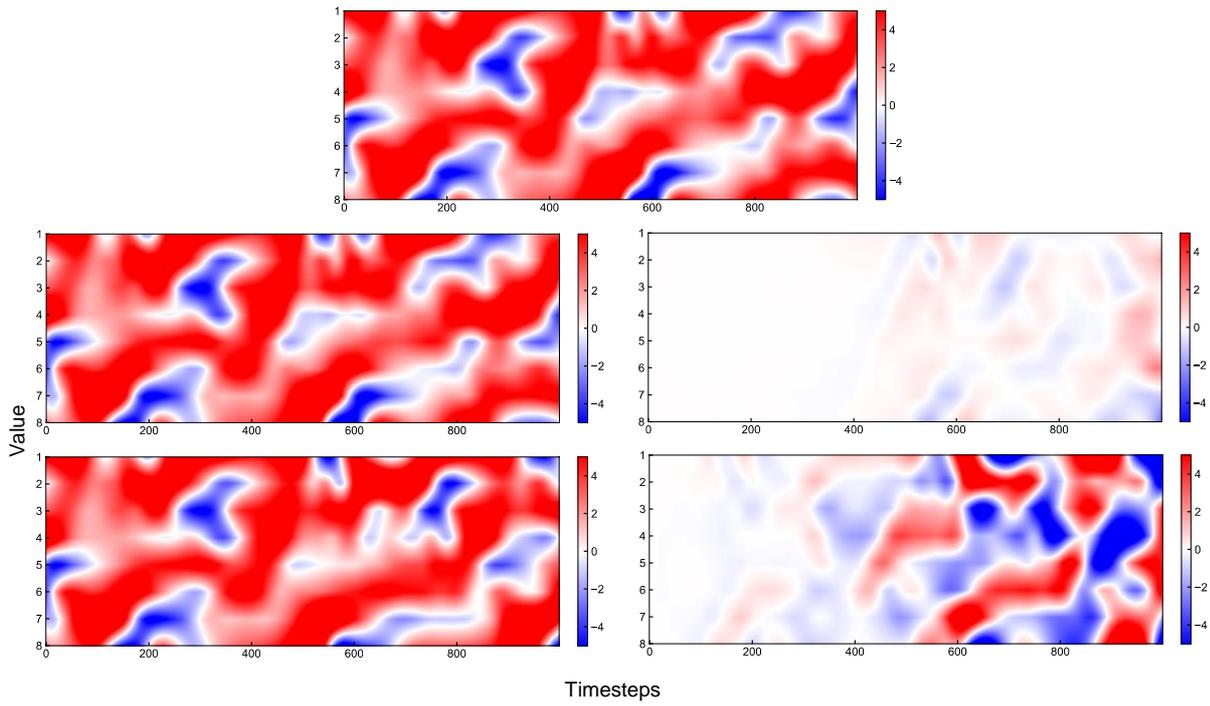

**Figure 2.** The Hovmöller diagram of (a) Nature Run, (b) Prediction of LETKF-Ext, (c) difference of (a) and (b), (d) Prediction of RC-Obs and (e) difference of (a) and (d). Horizontal axis shows the timesteps and vertical axis shows the nodal number. Value at each timestep and node is represented by the color.

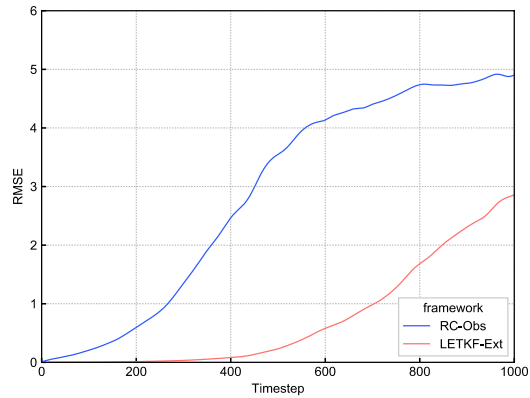

**Figure 3**. The *mRMSE* time series of the predictions of LETKF-Ext(red) and RC-Obs(blue) with perfect observation. Horizontal axis shows the timestep and vertical shows the value of *mRMSE*.

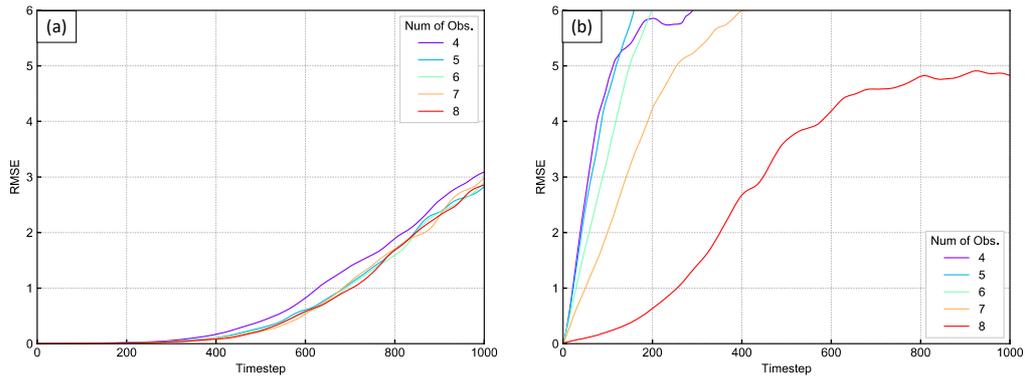

**Figure 4**. The *mRMSE* time series of the predictions of (a)LETKF-Ext and (b)RC-Obs with spatially sparse observation. Each color corresponds to the number of the observation points.

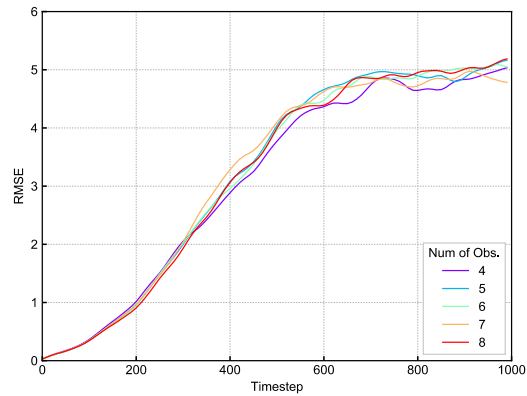

**Figure 5**. The same as figure4, for the RC-Anl prediction.

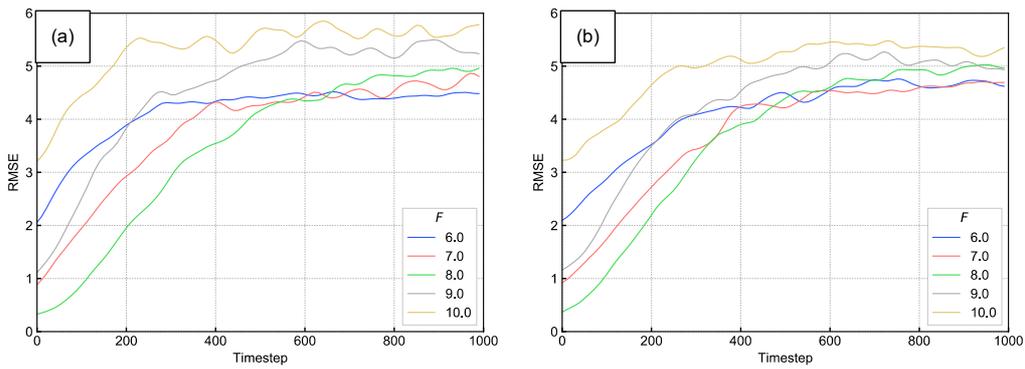

**Figure 6**. The *mRMSE* time series of the predictions of (a)LETKF-Ext and (b)RC-Anl with biased model. Each color corresponds to each value of *F* term.

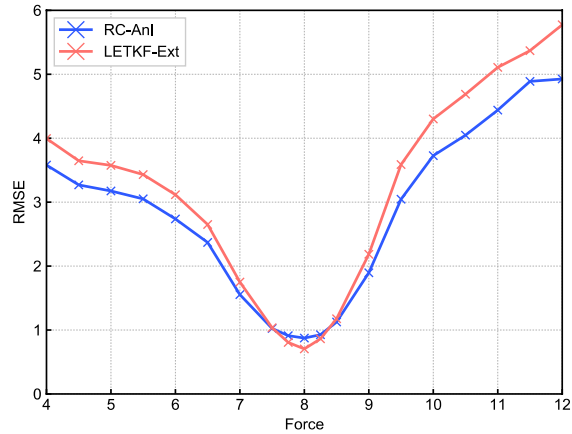

**Figure 7**. The *mRMSE(80)* of the predictions of LETKF-Ext(red) and RC-Anl(blue) for each model bias. Horizontal axis shows the value of the force parameter of equation (1) (8 is the true value) and vertical axis shows the value of *mRMSE*.

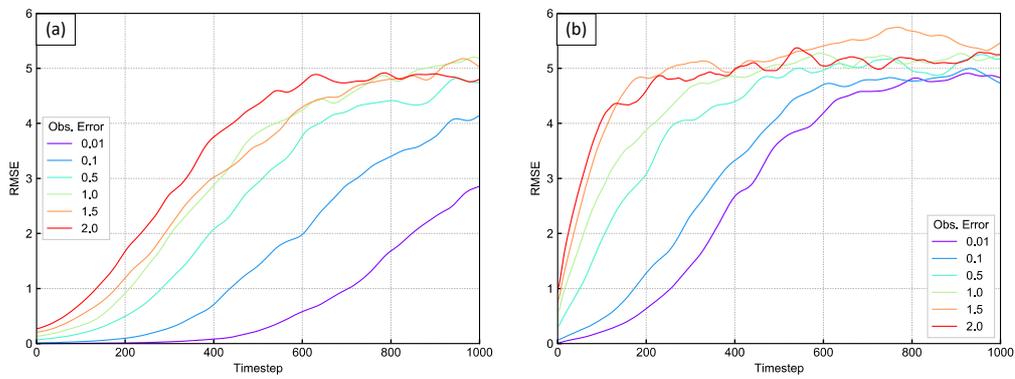

**Figure S1**. The *mRMSE* time series of the predictions of (a)LETKF-Ext and (b)RC-Obs noisy observation. Each color corresponds to the value of observation error parameter $e$.

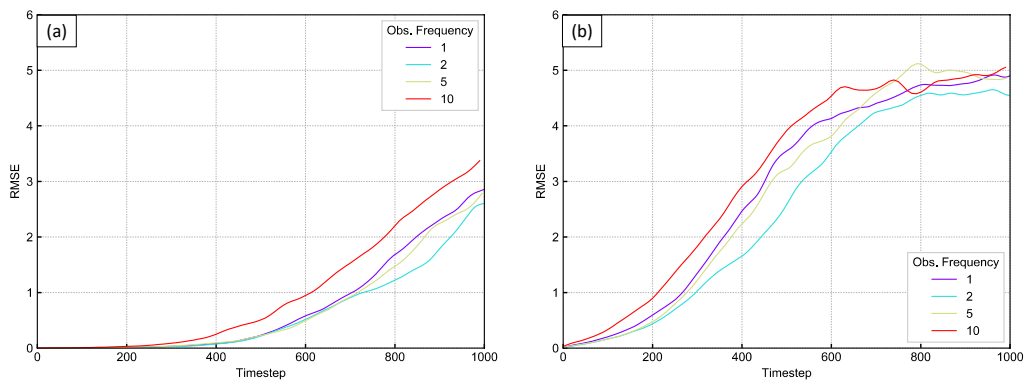

**Figure S2**. The *mRMSE* time series of the predictions of (a)LETKF-Ext and (b)RC-Obs with temporary sparse observation. Each color corresponds to the value of observation frequency. (If observation frequency is set to n, then we use the observation once in n timesteps.)

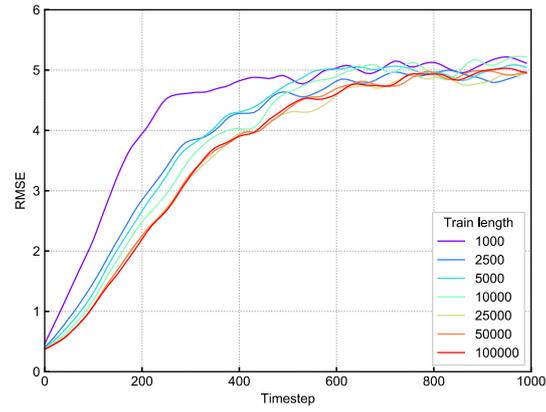

**Figure S3**. The *mRMSE* time series of the predictions of RC-Anl with various length of training data. Each color corresponds to the value of the size of training data.